\pdfoutput=1
\documentclass{article}
\PassOptionsToPackage{numbers,sort&compress}{natbib}
\usepackage[preprint]{neurips_2026}
\makeatletter
\renewcommand{\@noticestring}{Accepted at the Foundation Models for Temporal Systems Workshop at NeurIPS 2026.}
\makeatother

\usepackage[utf8]{inputenc}
\usepackage[T1]{fontenc}
\usepackage{hyperref}
\usepackage{url}
\usepackage{booktabs}
\usepackage{amsfonts,amsmath}
\usepackage[expansion=false]{microtype}
\usepackage{xcolor}
\usepackage{graphicx}
\usepackage{multirow}
\usepackage{caption}
\usepackage{subcaption}

\newcommand{\method}{\textsc{Kaleido}}
\newif\ifappendixfull\appendixfullfalse

\renewcommand{\arraystretch}{0.96}

\begin{document}
\title{\method{}: Input-Space Adaptation of a Vision Model\\ for Time-Series Forecasting\\ Through Gated Fold Geometries}

\author{%
  Xiangyu Shi$^{1}$\quad Qinghua Liu$^{2}$\quad Sam Heshmati$^{3}$\quad Zubin Abraham$^{3}$\\
  \normalfont\small $^{1}$Northwestern University\quad $^{2}$Ohio State University\quad $^{3}$Bosch Research North America\\
  \normalfont\small\texttt{xiangyushi2029@u.northwestern.edu}
}

\maketitle

\begin{abstract}
Time-series foundation models buy zero-shot forecasting with large temporal corpora; a vision model
needs none, since a natural image implicitly embeds the patterns a forecaster must model, and an
ImageNet-pretrained masked autoencoder forecasts a series by inpainting a rendering of it. A
rendered series is not a natural image, however, and closing that gap takes temporal-aware
adaptation. We show that the \emph{rendering geometry} --- how the series is folded and drawn ---
is a controllable, mixable axis for it. \method{} detects the dominant periods, renders a rule-generated
set of fold geometries, combines the inpaintings with a convex per-position gate fit on validation
only, and fuses the result with the zero-shot output at one fixed share, with no per-dataset
hyperparameter beyond the baseline's published settings. Training only LayerNorm
(0.05\%), \method{} lowers MSE by 13\% against the published zero-shot baseline on LTSF and,
frozen, by 6.6\%; on GIFT-Eval it improves the baseline by 7.4\% in MASE and 19.3\% in CRPS.
\end{abstract}

\section{Introduction}
\label{sec:intro}
Foundation models have reshaped time-series forecasting: pretrained on large temporal corpora,
they forecast unseen series zero-shot \citep{chronos,moirai}. Those corpora are costly to build and
hard to keep clear of the benchmarks. Language and vision foundation models, meanwhile, have grown
powerful in their own modalities; we ask whether that power transfers to time series, and how to
make it transfer better. Vision is the better source: a time series is closer to an image than to
text \citep{visionts}, both are continuous, a natural image implicitly embeds
the trends, periodic textures and abrupt changes a forecaster must model, and masked-autoencoder
inpainting \citep{mae} is itself forecasting in pixel space, whereas language-model routes
\citep{gpt4ts,chronos} must first discretize values, coarsening the continuous structure they
carry. VisionTS \citep{visionts} showed the route works: fold a normalized context window at its
seasonal period, render it as a grayscale image with the forecast region masked, and let a frozen
ImageNet MAE inpaint the missing pixels, read back as a competitive zero-shot forecast. We keep that backbone and take VisionTS under its published rendering as our baseline.

However, a rendered series still has a measurable domain gap to natural images: the same model reads
the same series differently depending on how it is drawn, yet the baseline fixes one hand-set
geometry per dataset. Closing this gap takes temporal-aware adaptation, on the \emph{input} (how the series is drawn)
or on the \emph{backbone} (its modules or computation). SSDA \citep{ssda}
does both; DMMV \citep{dmmv} renders only the seasonal part of a decomposition and adds modules
around the backbone. This paper takes the input route and asks what the backbone reads best.
\method{} detects the dominant periods, renders the series in a rule-generated set of fold
geometries (\emph{arms}), inpaints each with one pretrained backbone, and combines the decoded
forecasts with a convex per-position gate. No arm wins everywhere; the gate, a constrained
least-squares fit on validation, takes each arm where it is strong, so the
combination beats every single geometry; a never-finetuned zero-shot output fused at one fixed
share bounds the loss when the fit misleads. Nothing \method{} adds touches the backbone: gate and fusion act in
prediction space, and the trained tier finetunes only LayerNorm.

\paragraph{Contributions.} (i) We show that rendering geometry is a mixable expert axis for
adapting a pretrained vision backbone to time-series forecasting. (ii) We propose \method{}, an
automatic pipeline on that axis --- rule-based period detection, a generated arm set, a
validation-fit convex gate, fixed-share fusion with a never-fitted anchor --- with only LayerNorm
trained (0.05\%) and no per-dataset setting of its own. (iii) On LTSF the trained tier beats the baseline's
published zero-shot result on 20 of 24 cells (mean $-13\%$), with the best average and rank of its
group; frozen, it gains $6.6\%$. On all 97 GIFT-Eval configurations it improves the baseline by
$7.4\%$ in MASE and $19.3\%$ in CRPS.

\paragraph{Related work.} Period-folding architectures reshape a series at detected periods into 2D
representations but train task-specific networks from scratch \citep{timesnet,pdf,times2d}; we
reuse a pretrained backbone and make the fold itself the design space. The closest system, DMMV
\citep{dmmv}, renders only the seasonal component of a decomposition and fuses it with a numeric
trend branch through a per-series gate; our views are geometries of the whole series, our gate
varies with horizon position, and our anchor is never fitted. Combining forecasts is classical
\citep{bates1969}; our contribution is what we combine, not how.

\section{Method}
\label{sec:method}

\begin{figure}[t]
  \centering
  \includegraphics[width=\textwidth]{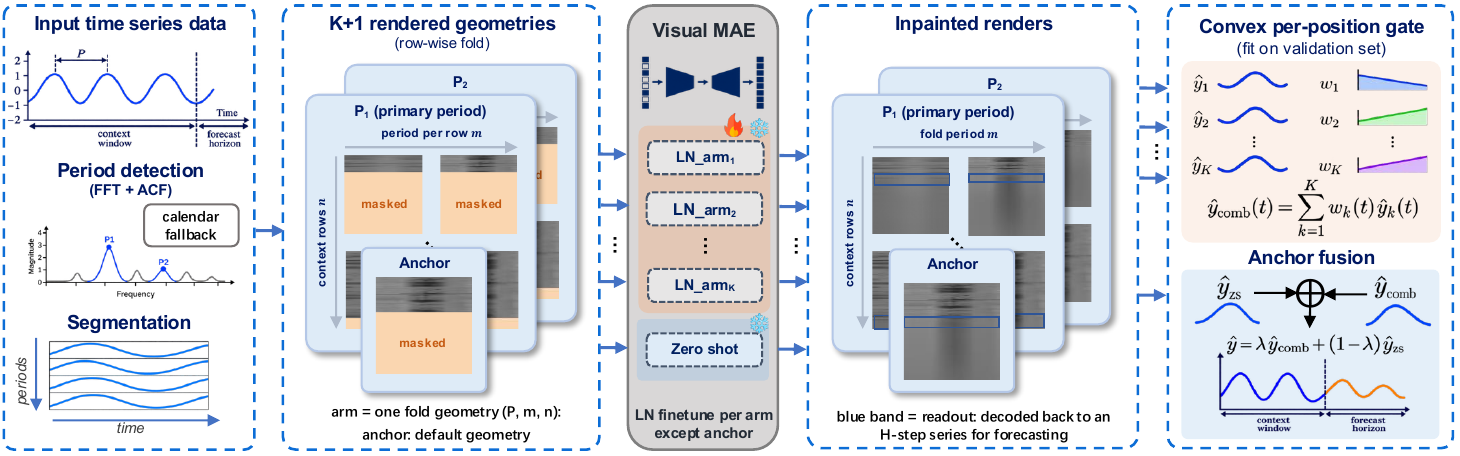}
  \caption{\method{} overview, left to right. Detected periods generate the arm set: one plane per
period, spanning periods per row $m$ and context rows $n$, for $K$ arms, all inpainted by one
shared MAE. The arms are frozen (zero-training tier) or carry their own LayerNorm parameters
(trained tier); the anchor, the backbone at the baseline's default geometry, is never finetuned and
never enters any fit. A convex per-position gate fit on validation windows combines the $K$ arms;
the anchor bypasses it and fuses in at $\lambda{=}0.75$.}
  \label{fig:overview}
\end{figure}

\paragraph{Preliminaries.} Following \citet{visionts}, a context window of length $L$ is
instance-normalized and folded at period $P$ into an $\lceil L/P\rceil\times P$ matrix; the matrix is
bilinearly resized onto a $224{\times}224$ grayscale canvas whose trailing rows (the horizon $H$) are
masked, and a frozen MAE ViT-B inpaints them, the band adjacent to the context boundary decoding
back into a forecast. We use that renderer unchanged and vary only the layout, drawing the fold
row-wise (the baseline renders the transpose, which moves the aggregate under $1\%$ on both benchmarks,
App.~\ref{app:orientation}). We call the layout the render's \emph{geometry}: how much time one
image row spans and how tightly the context is stacked into rows.

\paragraph{Arms: a rule-generated set of geometries.} An arm is one render, fixed by a period $P$,
a fold multiple $m$ and a row count $n$. Periods come from a rule-based detector --- FFT peaks
under thresholds frozen once, validated by autocorrelation, with a fallback to the calendar period
of the sampling frequency when it abstains --- and every confirmed period gets its own arms. $m$ sets the
horizontal compression: one image row holds $mP$ samples, so a smaller $m$ shows each cycle at
higher resolution. $n$ sets the
vertical compression: the folded context is squeezed into $n$ of the canvas's 14 patch rows and the
rest is masked. The baseline fixes $n$ indirectly through an alignment constant, so its look
changes with the dataset; we set $n$ directly on a log-spaced ladder, so every dataset spans the
same visual range. The \emph{anchor} is the baseline's own geometry rendered by our renderer: one
period per row ($m{=}1$) at its default alignment constant of 0.4. The detector's thresholds and the $n$ ladder are specified in
App.~\ref{app:rule}.

\paragraph{Gate: combining the arms.} Each arm shows the backbone a different view, and no arm
wins everywhere: the best one changes from cell to cell and can shift between validation and test,
so we combine rather than select. An arm can also be accurate early in the horizon and drift late
while another does the opposite, so the combination is fit \emph{per horizon position},
$\hat y_{\mathrm{comb}}(t)=\sum_k w_k(t)\,\hat y_k(t)$, which lets it take each arm where that arm
is strong and thereby exceed the best single arm. We adopt the simplest
convex family with this property: weights non-negative, summing to one at
every $t$ and linear across the horizon --- two coefficients per arm (first and last horizon positions), fit by constrained least
squares on the validation split only (formula in App.~\ref{app:rule}).

\paragraph{Anchor: bounding the fit.} The gate is only as good as its validation fit, so the final
forecast fuses it with the \emph{anchor}, the backbone's zero-shot output at the baseline's default
geometry, which never enters any fit:
$\hat y=\lambda\,\hat y_{\mathrm{comb}}+(1-\lambda)\,\hat y_{\mathrm{zs}}$ with one constant,
$\lambda{=}0.75$, in both tiers and on every dataset. Anchor and gate have different error sources: the
former is a fixed render that never adapts, the latter a fit that test-time drift can invalidate;
fusing them \citep{bates1969} bounds the loss where the fit misleads, at a small cost where the arms win.

\section{Experiments}
\label{sec:exp}

\paragraph{Setup.} \method{} runs one configuration everywhere: context rows
$n\in\{1,3,6,\mathrm{cap}\}$ and $m\in\{1,\tfrac12\}$ periods per row; $m\in\{2,4\}$ are tried but pruned
(App.~\ref{app:multiples}). Two tiers: the \emph{zero-training} tier
(\method{}-zs: frozen backbone) and the \emph{trained} tier, a 3-epoch LayerNorm finetune per
arm. Every render inherits the baseline's published context,
so both see the same input window (App.~\ref{app:protocol}).

\subsection{Long-term forecasting}
\label{sec:ltsf}

We use the six datasets the baseline publishes, under the standard chronological splits, scored by
multivariate MSE at horizons 96--720. Baselines (Table~\ref{tab:main}) are quoted as published,
at their own lookbacks \citep{visionts,visiontspp,moirai,chronos,dmmv,patchtst}; our own run of
the baseline's renderer at that configuration reproduces its published numbers to $0.01\%$ on
average over the 24 cells (App.~\ref{app:repro-baseline}).

The trained tier has the best average MSE (0.275 vs.\ 0.284 for
the closest baseline) and the best average rank of its group across the six datasets, and beats the
baseline's published zero-shot result on 20 of 24 cells (mean $-13\%$, up to $-35\%$ on Weather-96
and $-28\%$ averaged over ECL). All four losses are long-horizon ETT cells --- ETTh1-720 and ETTh2 at 192, 336 and 720
--- the family where a validation fit transfers worst to test (App.~\ref{app:drift}); fusing
with the anchor keeps each of them below $+3\%$. The zero-training tier, the same arms and gate on
the frozen backbone, improves the baseline by $6.6\%$ per cell and matches the
best average rank in its group, held by VisionTS++, whose backbone is continually pretrained on
large-scale time series while ours sees none.

\begin{table}[!ht]
  \caption{LTSF, multivariate MSE averaged over horizons; the trained column is the mean over 3
  seeds (ECL: 1). Groups: no backbone parameter trained on the target dataset (left) versus trained on it
  (right). \textbf{Bold}: best in group, co-bolded within 0.5\%. Per-cell values in App.~\ref{app:master}.}
  \label{tab:main}
  \centering\footnotesize
  \setlength{\tabcolsep}{2.2pt}
  \begin{tabular}{lccccc|cccc}
\toprule
 & \multicolumn{5}{c|}{\textbf{Zero-shot}} & \multicolumn{4}{c}{\textbf{Fine-tuned / supervised}} \\
\cmidrule(lr){2-6}\cmidrule(lr){7-10}
 & \textbf{\method-zs} & VisionTS-zs & VisionTS++ & Moirai\textsubscript{B} & Chronos\textsubscript{L} & \textbf{\method} & VisionTS-fs & DMMV-a & PatchTST \\
\midrule
ETTh1 & \textbf{0.384} & 0.389 & 0.402 & 0.434 & 0.589 & \textbf{0.376} & 0.395 & 0.395 & 0.413 \\
ETTh2 & \textbf{0.330} & 0.333 & 0.333 & 0.346 & 0.455 & 0.338 & 0.336 & 0.337 & \textbf{0.330} \\
ETTm1 & \textbf{0.349} & 0.373 & 0.360 & 0.382 & 0.556 & \textbf{0.320} & 0.338 & 0.340 & 0.351 \\
ETTm2 & 0.256 & 0.281 & \textbf{0.244} & 0.272 & 0.295 & \textbf{0.242} & 0.261 & 0.256 & 0.255 \\
Weather & 0.243 & 0.269 & \textbf{0.222} & 0.238 & 0.279 & 0.222 & 0.227 & \textbf{0.217} & 0.225 \\
ECL & \textbf{0.184} & 0.207 & \textbf{0.183} & 0.188 & 0.204 & \textbf{0.150} & 0.156 & 0.158 & 0.159 \\
\midrule
Avg.\ MSE & \textbf{0.291} & 0.309 & \textbf{0.291} & 0.310 & 0.396 & \textbf{0.275} & 0.285 & 0.284 & 0.289 \\
Avg.\ rank & \textbf{1.67} & 3.50 & \textbf{1.67} & 3.33 & 4.83 & \textbf{1.67} & 2.67 & 2.67 & 3.00 \\
$1^{\mathrm{st}}$ count & 4 & 0 & 3 & 0 & 0 & 4 & 0 & 1 & 1 \\
\bottomrule
\end{tabular}

\end{table}

\subsection{Transfer to GIFT-Eval}
\label{sec:gift}

Both tiers ran unchanged --- same detector, arm rule, gate and $\lambda$ --- on all 97 GIFT-Eval
configurations \citep{gifteval}, scored by the official evaluator. Baselines are the benchmark's
published result files; for the one we adapt we re-run it at its published configuration, so
backbone, data and evaluator match on both sides ($\star$); its leaderboard entry has no public
replication code.

\begin{table}[!ht]
  \caption{GIFT-Eval, the benchmark's own aggregation (MASE and CRPS normalised by Seasonal-Naive,
  geometric mean over 97 configurations). Per-configuration values in App.~\ref{app:gift-master}.}
  \label{tab:gift}
  \centering\footnotesize
  \setlength{\tabcolsep}{1.6pt}
  \begin{tabular}{lcccccccccc}
\toprule
 & TimesFM-2 & C-Bolt-B & C-Bolt-S & \textbf{\method{}} & PatchTST & Chronos-L & \textbf{\method{}-zs} & iTrans. & VisionTS$^{\star}$ & Moirai-B \\
\midrule
nMASE & 0.758 & 0.808 & 0.822 & \textbf{0.832} & 0.849 & 0.870 & \textbf{0.879} & 0.893 & 0.899 & 0.901 \\
nCRPS & 0.550 & 0.574 & 0.577 & \textbf{0.634} & 0.587 & 0.647 & \textbf{0.671} & 0.620 & 0.786 & 0.610 \\
leakage & \checkmark & \checkmark & \checkmark & -- & -- & \checkmark & -- & -- & -- & -- \\
\bottomrule
\end{tabular}

\end{table}

The trained tier improves on that baseline by $7.4\%$ in MASE and $19.3\%$ in CRPS, winning on 74
and 91 of 97; among the unflagged entries it has the best MASE, while three keep a lower CRPS.
The CRPS margin over the baseline is larger because the arms induce a predictive distribution, submitted as nine
quantiles around the fused point (App.~\ref{app:gift-master}); a single render submits a point,
whose CRPS is its absolute error. The zero-training tier, backbone frozen, gains $2.2\%$ and
$14.7\%$, winning on 60 and 86 and beating Seasonal-Naive on 76.

\subsection{Ablation study}
\label{sec:ablate}
\begin{figure}[!ht]
  \begin{subfigure}[t]{0.32\linewidth}
    \includegraphics[width=\linewidth]{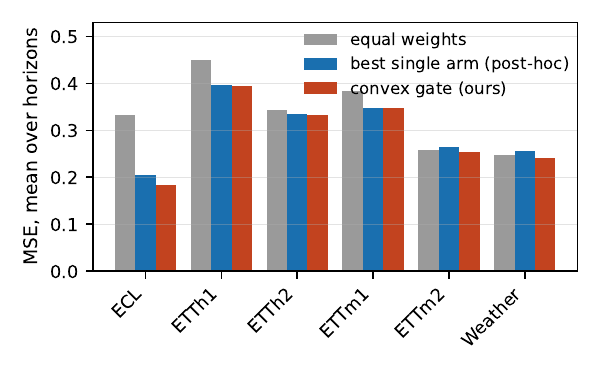}
    \caption{Gate vs.\ its alternatives.}
  \end{subfigure}\hfill
  \begin{subfigure}[t]{0.32\linewidth}
    \includegraphics[width=\linewidth]{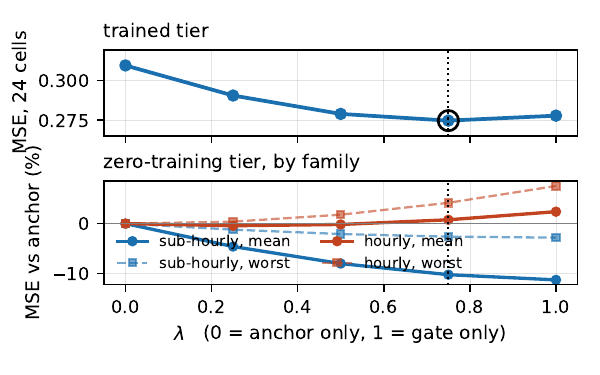}
    \caption{Anchor share $\lambda$.}
  \end{subfigure}\hfill
  \begin{subfigure}[t]{0.32\linewidth}
    \includegraphics[width=\linewidth]{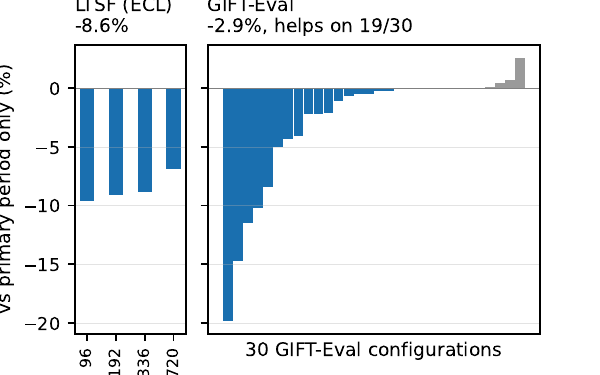}
    \caption{Multi-period detection.}
  \end{subfigure}
  \caption{Ablations. (a) MSE per dataset at the combination stage, no anchor, zero-training tier;
  (b) the fusion share, trained-tier MSE over the 24 cells of Table~\ref{tab:main} (top) and the
  zero-training change against the anchor by frequency (bottom); (c) primary period versus all detected periods.}
  \label{fig:ablate}
\end{figure}

\paragraph{Is the gate needed? (Fig.~\ref{fig:ablate}a)} We compare the convex gate with
equal weights and the post-hoc best single arm. Equal weights is worse on every dataset and on 23 of 24
cells, collapsing wherever the arms differ widely in accuracy (ECL $+71$ to $+97\%$), because a junk arm
enters the average at full weight. Even the post-hoc best single arm, which no protocol can select,
is beaten on 5 of 6 datasets and by $+4.2\%$ over cells (App.~\ref{app:combiner}).

\paragraph{How much anchor? (Fig.~\ref{fig:ablate}b)} In the zero-training tier, the two families pull in
opposite directions: the sub-hourly cells improve monotonically towards the gate
($-11.3\%$ at $\lambda{=}1$) while the hourly cells need the anchor, the worst $+7.6\%$ there. Over the 24 cells of Table~\ref{tab:main} the trained tier's mean MSE is flat around the
operating point and rises towards the anchor: 0.310, 0.291, 0.279, 0.275, 0.278 at
$\lambda{=}0,\ldots,1$. Choosing $\lambda$ per cell on
validation is falsified: the gate is fit on those windows, so validation prefers
$\lambda{=}1$ on 22 of 24 cells and costs up to $+5.1\%$ on the drifting ones
(App.~\ref{app:lambda}).

\paragraph{Does multi-period detection pay? (Fig.~\ref{fig:ablate}c)} Adding the arms of the second
confirmed period, the rest unchanged, lowers MSE at the combination stage by $8.6\%$ on the four LTSF cells that have
one and MASE by $2.9\%$ over the 30 GIFT-Eval configurations that render one, helping on 19
(App.~\ref{app:multiperiod}).

\section{Conclusion}
\label{sec:conclusion}
A pretrained vision model reads a time series through whatever geometry the render gives it, and
that geometry is a design space, not fixed preprocessing. Folding the series several ways by rule,
inpainting each with one backbone and mixing the results with a convex per-position gate beats both
the equal-weight average and the best single geometry. With only LayerNorm trained (0.05\%),
\method{} lowers MSE by 13\% on 20 of 24 LTSF cells and improves the baseline by 7.4\% on
GIFT-Eval; frozen, it gains 6.6\%. Future extensions include fusing the arms
hierarchically along the different geometry axes, designing a more capable gate, and letting the anchor share adapt.

\bibliographystyle{plainnat}
\bibliography{references}

\newpage
\appendix
\section{Render orientation}
\label{app:orientation}
The fold has an orientation: we draw each cycle as an image \emph{row} (time flows top to bottom),
the baseline's renderer draws the transpose. Its own orientation ablation (Table~18 there) flips
the image horizontally or vertically, transformations that preserve which axis is time, and
concludes reconstruction is isotropic; a transpose \emph{swaps} the axes and is not covered by that
test. The choice is not free but it is small and sign-mixed. On GIFT-Eval, rendering all 97
configurations both ways (identical pipeline, image and patch mask transposed at the MAE boundary)
moves single configurations by $-16.1\%$ to $+16.8\%$ MASE and the aggregate by $+0.6\%$ (0.892
row-wise vs 0.886 column-wise, CRPS 0.678 vs 0.680), both before the repair of
App.~\ref{app:gift-master} so the pair is on identical code. The trained tier, re-trained
column-wise on all 97 (this pair predates the move to the global $\lambda{=}0.75$ and is reported
as run, under the earlier frequency rule), gives 0.843 vs 0.840 nMASE and 0.636 vs 0.636 CRPS, with
eight configurations beyond $\pm5\%$.

On LTSF the column-wise runs cover the five ETT and Weather datasets plus ILI, 24 cells in all;
ECL was never re-rendered column-wise. The zero-training tier (same arms, contexts, anchor and
gate) moves single cells by $-2.5\%$ to $+5.7\%$ and their mean by $+0.8\%$:
\begin{center}\small\begin{tabular}{lrrrr}
\toprule
cell & row-wise & column-wise & $\Delta$ & backbone \\
\midrule
ETTh1-96 & 0.3482 & 0.3433 & -1.4\% & 0.3431 \\
ETTh1-192 & 0.3792 & 0.3824 & +0.8\% & 0.3693 \\
ETTh1-336 & 0.3972 & 0.3977 & +0.1\% & 0.3812 \\
ETTh1-720 & 0.4122 & 0.4019 & -2.5\% & 0.4094 \\
ETTh2-96 & 0.2695 & 0.2680 & -0.6\% & 0.2735 \\
ETTh2-192 & 0.3285 & 0.3305 & +0.6\% & 0.3322 \\
ETTh2-336 & 0.3514 & 0.3633 & +3.4\% & 0.3540 \\
ETTh2-720 & 0.3712 & 0.3925 & +5.7\% & 0.3695 \\
ETTm1-96 & 0.3135 & 0.3165 & +0.9\% & 0.3403 \\
ETTm1-192 & 0.3331 & 0.3357 & +0.8\% & 0.3623 \\
ETTm1-336 & 0.3498 & 0.3530 & +0.9\% & 0.3772 \\
ETTm1-720 & 0.3985 & 0.3993 & +0.2\% & 0.4225 \\
ETTm2-96 & 0.1836 & 0.1863 & +1.5\% & 0.2299 \\
ETTm2-192 & 0.2276 & 0.2296 & +0.9\% & 0.2659 \\
ETTm2-336 & 0.2703 & 0.2736 & +1.2\% & 0.2986 \\
ETTm2-720 & 0.3435 & 0.3435 & +0.0\% & 0.3527 \\
Weather-96 & 0.1898 & 0.1892 & -0.3\% & 0.2219 \\
Weather-192 & 0.2199 & 0.2201 & +0.1\% & 0.2436 \\
Weather-336 & 0.2552 & 0.2594 & +1.6\% & 0.2761 \\
Weather-720 & 0.3082 & 0.3154 & +2.3\% & 0.3203 \\
ILI-24 & 2.4944 & 2.5188 & +1.0\% & 2.4834 \\
ILI-36 & 2.3303 & 2.3588 & +1.2\% & 2.2319 \\
ILI-48 & 2.2631 & 2.2985 & +1.6\% & 2.1043 \\
ILI-60 & 2.1594 & 2.1448 & -0.7\% & 1.9373 \\
\bottomrule
\end{tabular}
\end{center}
\noindent The trained tier, every LayerNorm finetune redone on the transposed render, stays within
$\pm1.8\%$ per cell except ETTh2-720 ($-4.8\%$) and ILI, whose tiny validation split makes the gate
itself noisy, for a mean of $-0.3\%$ over the same 24 cells ($0.0\%$ without ILI). Superscripts
give the seed count.
\begin{center}\scriptsize\begin{tabular}{lrrr}
\toprule
cell & row-wise & column-wise & $\Delta$ \\
\midrule
ETTh1-96 & 0.3357 & 0.3346$^{(3)}$ & -0.3\% \\
ETTh1-192 & 0.3675 & 0.3686$^{(3)}$ & +0.3\% \\
ETTh1-336 & 0.3836 & 0.3838$^{(3)}$ & +0.1\% \\
ETTh1-720 & 0.4158 & 0.4231$^{(3)}$ & +1.8\% \\
ETTh2-96 & 0.2705 & 0.2674$^{(3)}$ & -1.1\% \\
ETTh2-192 & 0.3356 & 0.3338$^{(3)}$ & -0.5\% \\
ETTh2-336 & 0.3545 & 0.3538$^{(3)}$ & -0.2\% \\
ETTh2-720 & 0.3934 & 0.3743$^{(3)}$ & -4.8\% \\
ETTm1-96 & 0.2693 & 0.2741$^{(2)}$ & +1.8\% \\
ETTm1-192 & 0.3069 & 0.3094$^{(2)}$ & +0.8\% \\
ETTm1-336 & 0.3304 & 0.3344$^{(2)}$ & +1.2\% \\
ETTm1-720 & 0.3731 & 0.3745$^{(2)}$ & +0.4\% \\
ETTm2-96 & 0.1596 & 0.1596$^{(2)}$ & -0.0\% \\
ETTm2-192 & 0.2118 & 0.2103$^{(2)}$ & -0.7\% \\
ETTm2-336 & 0.2593 & 0.2623$^{(2)}$ & +1.2\% \\
ETTm2-720 & 0.3383 & 0.3405$^{(2)}$ & +0.7\% \\
Weather-96 & 0.1439 & 0.1446$^{(3)}$ & +0.5\% \\
Weather-192 & 0.1868 & 0.1878$^{(3)}$ & +0.5\% \\
Weather-336 & 0.2437 & 0.2426$^{(3)}$ & -0.4\% \\
Weather-720 & 0.3146 & 0.3100$^{(2)}$ & -1.5\% \\
ILI-24 & 2.1665 & 1.9799$^{(3)}$ & -8.6\% \\
ILI-36 & 1.8640 & 1.9491$^{(3)}$ & +4.6\% \\
ILI-48 & 1.8358 & 1.8662$^{(3)}$ & +1.7\% \\
ILI-60 & 1.7744 & 1.6843$^{(3)}$ & -5.1\% \\
\bottomrule
\end{tabular}
\end{center}
\noindent We standardize on row-wise everywhere and keep the column-wise runs as the ablation. On
LTSF the baseline column reports its published numbers, which come from its column-wise pipeline,
while our anchor is our row-wise render of the same configuration; on GIFT-Eval the baseline is its
own submission. Each pipeline is internally consistent and the fusion acts in prediction space, not
pixel space.

\section{Period detector, arm rule and gate}
\label{app:rule}
\paragraph{Period detection.} FFT peak detection with thresholds fixed once, before any reported
cell was run:
a candidate must be a local maximum of the channel-averaged spectrum, at least $2.5\times$ the median
of its $\pm15\%$ spectral neighbourhood (sharpness), at least $5\times$ the global median and 9\% of the
top peak (amplitude), and fit at least twice into the context; harmonics of a kept period are
suppressed. The LTSF period table was produced by these gates alone. The final detector, used on
GIFT-Eval, additionally detrends, requires at least three cycles in the training series, validates
every candidate as a local maximum of the autocorrelation at the candidate lag (ACF $\ge 0.05$), and
admits a calendar period that passes every gate except amplitude when its ACF support is
$\ge 0.10$; it reproduces the LTSF period table exactly. When it abstains, the fold period is the standard
calendar seasonality of the sampling frequency (the same freq$\to$period map the benchmarks' MASE
scaling uses); on GIFT-Eval the anchor folds at the detected primary period, which is this calendar period
when the detector abstains.
Table~\ref{tab:detector} lists every detection against its reference.
\begin{table}[h]
  \caption{Every detected period against the reference seasonality of its dataset.}
  \centering\tiny\setlength{\tabcolsep}{3pt}
  \begin{tabular}{lll}
\toprule
\multicolumn{3}{l}{\textbf{LTSF} (detector v1 = v2, frozen; reference = community seasonality)}\\
dataset & detected & reference \\ \midrule
ETTh1 & 24 & 24 \\
ETTh2 & 24 & 24 \\
ETTm1 & 96 & 96 \\
ETTm2 & 96 & 96 \\
Weather & 144 & 144 \\
ECL & 24, 167 & 24, 168 \\
\bottomrule
\end{tabular}

\begin{tabular}{llll}
\toprule
\multicolumn{4}{l}{\textbf{GIFT-Eval} (detector v2; reference = calendar seasonality of the sampling frequency;}\\
\multicolumn{4}{l}{`---' = nothing detected $\Rightarrow$ calendar fallback at render time; several M4/daily panels are weakly seasonal by construction)}\\
dataset & freq & calendar ref.\ & detected (majority vote) \\ \midrule
bitbrains\_fast\_storage/5T & 5T & 288, 2016 & --- \\
bitbrains\_fast\_storage/H & H & 24, 168 & 24 \\
bitbrains\_rnd/5T & 5T & 288, 2016 & --- \\
bitbrains\_rnd/H & H & 24, 168 & --- \\
bizitobs\_application & 10S & 6, 60, 360, 8640 & 353, 2945 \\
bizitobs\_l2c/5T & 5T & 288, 2016 & 1998 \\
bizitobs\_l2c/H & H & 24, 168 & 166 \\
bizitobs\_service & 10S & 6, 60, 360, 8640 & 353, 2945 \\
car\_parts\_with\_missing & M & 12 & --- \\
covid\_deaths & D & 7, 365 & --- \\
electricity/15T & 15T & 96, 672 & 96, 35072, 666 \\
electricity/D & D & 7, 365 & 365, 7 \\
electricity/H & H & 24, 168 & 24, 8768, 166 \\
electricity/W & W & 52 & 52 \\
ett1/15T & 15T & 96, 672 & 96, 11613 \\
ett1/D & D & 7, 365 & --- \\
ett1/H & H & 24, 168 & 24, 2903 \\
ett1/W & W & 52 & --- \\
ett2/15T & 15T & 96, 672 & 96 \\
ett2/D & D & 7, 365 & --- \\
ett2/H & H & 24, 168 & 24 \\
ett2/W & W & 52 & --- \\
hierarchical\_sales/D & D & 7, 365 & 7, 360 \\
hierarchical\_sales/W & W & 52 & 52 \\
hospital & M & 12 & --- \\
jena\_weather & 10T & 144, 1008 & --- \\
jena\_weather/10T & 10T & 144, 1008 & --- \\
jena\_weather/D & D & 7, 365 & --- \\
jena\_weather/H & H & 24, 168 & --- \\
kdd\_cup\_2018\_with\_missing/D & D & 7, 365 & --- \\
kdd\_cup\_2018\_with\_missing/H & H & 24, 168 & 24 \\
LOOP\_SEATTLE/5T & 5T & 288, 2016 & 2022, 288 \\
LOOP\_SEATTLE/D & D & 7, 365 & 7 \\
LOOP\_SEATTLE/H & H & 24, 168 & 168, 24 \\
m4\_daily & D & 7, 365 & --- \\
m4\_hourly & H & 24, 168 & 24, 168 \\
m4\_monthly & M & 12 & 12 \\
m4\_quarterly & Q & 4 & --- \\
m4\_weekly & W & 52 & --- \\
m4\_yearly & A & none & --- \\
M\_DENSE/D & D & 7, 365 & 7, 122, 182 \\
M\_DENSE/H & H & 24, 168 & 24, 168, 2920, 4380 \\
restaurant & D & 7, 365 & 7 \\
saugeenday/D & D & 7, 365 & 365, 7914, 4748 \\
saugeenday/M & M & 12 & 12 \\
saugeenday/W & W & 52 & 52, 1130 \\
solar/10T & 10T & 144, 1008 & 144 \\
solar/D & D & 7, 365 & --- \\
solar/H & H & 24, 168 & 24 \\
solar/W & W & 52 & --- \\
SZ\_TAXI/15T & 15T & 96, 672 & 96, 496, 48 \\
SZ\_TAXI/H & H & 24, 168 & 24 \\
temperature\_rain\_with\_missing & D & 7, 365 & --- \\
us\_births/D & D & 7, 365 & 7, 365, 1826 \\
us\_births/M & M & 12 & 12, 60 \\
us\_births/W & W & 52 & 52 \\
\bottomrule
\end{tabular}

  \label{tab:detector}\label{app:detector}
\end{table}

\paragraph{Arm rule.} With $R = 14L/(L+H)$ the geometric maximum of context rows, the ladder is
$n\in\{1,3,6,\lfloor R\rfloor\}$ with alignment constant $\alpha=(n+\tfrac12)/R$; the $m{=}\tfrac12$
ladder folds at the nearest integer to $P/2$ when $P$ is odd. Whole-multiple folds ($m\in\{2,4\}$)
are excluded ($\le$1.5\% on any cell, App.~\ref{app:multiples}). Arms whose fold needs more than the context
holds (fewer than two cycles) are dropped in the zero-training tier; the trained tier keeps them
because finetuning repairs them; the control is in App.~\ref{app:gift-master}.

\paragraph{Gate.} The per-position weights are
$w_k(t) = a_k\,(H-t)/(H-1) + b_k\,(t-1)/(H-1)$, fit by constrained least squares on the
validation split with $\sum_k a_k = \sum_k b_k = 1$ and $a,b \ge 0$: two coefficients per arm,
linear in the horizon position and convex at every $t$.

\paragraph{Out of distribution.} On GIFT-Eval's short, trend-dominated panels
the frozen detector missed known seasonality (M4 monthly/quarterly) and a spectral boundary
artifact could hijack the reference peak. Detrending, a minimum-cycles bound, autocorrelation
validation of every spectral candidate, and a calendar-frequency rescue repair the hijack cases
(2/2) and part of the misses (2/5) while leaving every LTSF detection unchanged; the remaining
weakly seasonal panels fall back to the calendar period by construction.

\section{Datasets, contexts and training protocol}
\label{app:protocol}
\paragraph{Datasets.} The six LTSF datasets the baseline publishes: ETTh1 and ETTh2 (hourly
electricity-transformer temperature), ETTm1 and ETTm2 (the same sources at 15 minutes), Weather
(10-minute meteorological readings) and ECL (hourly electricity load), under the standard
chronological splits.

\paragraph{Contexts and anchors.} On LTSF every render, ours and the backbone's alike, uses the
context length VisionTS publishes for the dataset (their Tables 8 and 19: full-shot contexts for
trained arms, zero-shot contexts for the zero-training tier), inherited unchanged so that backbone
and \method{} see the same input window. In the trained tier the anchor's context is chosen per
cell on validation between the published full-shot and zero-shot values (validation picks the
zero-shot context on ETTh1, Weather and ECL-720 and the full-shot context elsewhere; fixing it at
the published zero-shot context instead moves each of the 20 non-ECL cells by at most 1.8\% and
their mean MSE from 0.2997 to 0.2989, with the same 16 wins against the published zero-shot
baseline); in the zero-training
tier the anchor is the render at the published zero-shot configuration, the same configuration its
arms use. On GIFT-Eval every render runs at the single global context the backbone publishes for
the whole benchmark, with its default $r{=}c{=}0.4$.
\paragraph{Trained tier.} Per arm, 3 epochs, batch 32, LayerNorm parameters only (the scale and shift of every LayerNorm in encoder and decoder, 55{,}808 values, 0.05\% of
parameters), the loss taken in series space on the decoded forecast, seeds \{2021, 2022, 2023\}; the reported cell is the mean over
seeds of the fused forecast's MSE (the four ECL cells are single-seed; per-cell seed standard
deviation has median 0.4\% and maximum 1.2\% over the three-seed cells).
\paragraph{Baseline lookbacks.} Baseline numbers are as published and their lookbacks are not
matched to ours --- the standing caveat of the benchmark. DMMV fixes 336; the PatchTST column is the
per-cell best of lookbacks 336/512 as compiled by \citet{gpt4ts} and reused by VisionTS, which
favors the baseline; Chronos-L per-cell values are as reported in VisionTS++. All external numbers
were re-verified against the source PDFs; one transcription error inherited from a survey table
(PatchTST ECL-192) was corrected to the original 0.147.

\section{Whole-multiple folds}
\label{app:multiples}
\begin{center}\includegraphics[width=.98\linewidth]{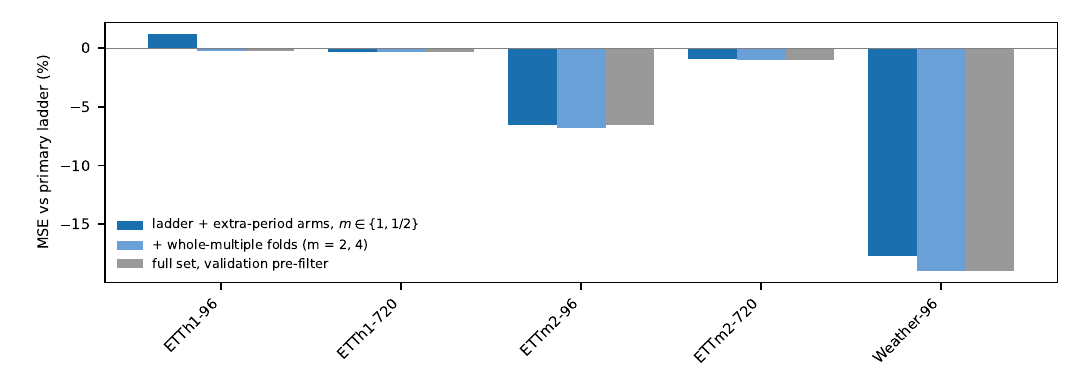}\end{center}
\noindent The exploratory sweep (five cells, $L{=}1152$): the arm
rule's set against the same set plus $m\in\{2,4\}$, relative to the primary ladder. The
whole-multiple folds move no cell by more than 1.5\% (mean $-0.6\%$) for twice the arm count, and
are excluded.

\section{Full per-cell LTSF results}
\label{app:master}
Every confirmed measurement, one row per (dataset, horizon) cell; the main table aggregates from
this record. Values within 0.5\% of the row best are co-bolded in the main table. VisionTS-zs is the
published number (VisionTS Table~17).
\begin{center}\tiny\setlength{\tabcolsep}{2.5pt}\begin{tabular}{llccc|cccccc}
\toprule
 & $H$ & \method-zs & \method{} (ft) & VisionTS-zs & VisionTS-fs & VisionTS++ & Moirai\textsubscript{B} & Chronos\textsubscript{L} & DMMV-a & PatchTST \\
\midrule
\multirow{4}{*}{ETTh1} & 96 & 0.3482 & 0.3357\,{\tiny$\pm$0.0006} & 0.3530 & 0.3470 & 0.3690 & 0.3840 & 0.4410 & 0.3540 & 0.3700 \\
 & 192 & 0.3792 & 0.3675\,{\tiny$\pm$0.0009} & 0.3920 & 0.3850 & 0.3990 & 0.4250 & 0.5020 & 0.3930 & 0.4130 \\
 & 336 & 0.3972 & 0.3836\,{\tiny$\pm$0.0007} & 0.4070 & 0.4070 & 0.4150 & 0.4560 & 0.5760 & 0.3870 & 0.4220 \\
 & 720 & 0.4122 & 0.4158\,{\tiny$\pm$0.0025} & 0.4060 & 0.4390 & 0.4240 & 0.4700 & 0.8350 & 0.4450 & 0.4470 \\
\midrule
\multirow{4}{*}{ETTh2} & 96 & 0.2695 & 0.2705\,{\tiny$\pm$0.0008} & 0.2710 & 0.2690 & 0.2770 & 0.2770 & 0.3200 & 0.2940 & 0.2740 \\
 & 192 & 0.3285 & 0.3356\,{\tiny$\pm$0.0014} & 0.3280 & 0.3320 & 0.3330 & 0.3400 & 0.4060 & 0.3390 & 0.3390 \\
 & 336 & 0.3514 & 0.3545\,{\tiny$\pm$0.0012} & 0.3450 & 0.3510 & 0.3500 & 0.3710 & 0.4920 & 0.3220 & 0.3290 \\
 & 720 & 0.3712 & 0.3934\,{\tiny$\pm$0.0015} & 0.3880 & 0.3900 & 0.3700 & 0.3940 & 0.6030 & 0.3920 & 0.3790 \\
\midrule
\multirow{4}{*}{ETTm1} & 96 & 0.3135 & 0.2693\,{\tiny$\pm$0.0009} & 0.3410 & 0.2810 & 0.3160 & 0.3350 & 0.4570 & 0.2790 & 0.2900 \\
 & 192 & 0.3331 & 0.3069\,{\tiny$\pm$0.0018} & 0.3600 & 0.3220 & 0.3470 & 0.3660 & 0.5300 & 0.3170 & 0.3320 \\
 & 336 & 0.3498 & 0.3304\,{\tiny$\pm$0.0012} & 0.3770 & 0.3560 & 0.3680 & 0.3910 & 0.5770 & 0.3510 & 0.3660 \\
 & 720 & 0.3985 & 0.3731\,{\tiny$\pm$0.0023} & 0.4160 & 0.3910 & 0.4080 & 0.4340 & 0.6600 & 0.4110 & 0.4160 \\
\midrule
\multirow{4}{*}{ETTm2} & 96 & 0.1836 & 0.1596\,{\tiny$\pm$0.0001} & 0.2280 & 0.1690 & 0.1690 & 0.1950 & 0.1970 & 0.1720 & 0.1650 \\
 & 192 & 0.2276 & 0.2118\,{\tiny$\pm$0.0005} & 0.2620 & 0.2250 & 0.2160 & 0.2470 & 0.2540 & 0.2270 & 0.2200 \\
 & 336 & 0.2703 & 0.2593\,{\tiny$\pm$0.0010} & 0.2930 & 0.2780 & 0.2600 & 0.2910 & 0.3130 & 0.2720 & 0.2740 \\
 & 720 & 0.3435 & 0.3383\,{\tiny$\pm$0.0011} & 0.3430 & 0.3720 & 0.3300 & 0.3550 & 0.4160 & 0.3510 & 0.3620 \\
\midrule
\multirow{4}{*}{Weather} & 96 & 0.1898 & 0.1439\,{\tiny$\pm$0.0009} & 0.2200 & 0.1420 & 0.1450 & 0.1670 & 0.1940 & 0.1430 & 0.1490 \\
 & 192 & 0.2199 & 0.1868\,{\tiny$\pm$0.0002} & 0.2440 & 0.1910 & 0.1870 & 0.2090 & 0.2490 & 0.1870 & 0.1940 \\
 & 336 & 0.2552 & 0.2437\,{\tiny$\pm$0.0016} & 0.2800 & 0.2460 & 0.2400 & 0.2560 & 0.3020 & 0.2370 & 0.2450 \\
 & 720 & 0.3082 & 0.3146\,{\tiny$\pm$0.0025} & 0.3300 & 0.3280 & 0.3170 & 0.3210 & 0.3720 & 0.3020 & 0.3140 \\
\midrule
\multirow{4}{*}{ECL} & 96 & 0.1536 & 0.1233 & 0.1770 & 0.1260 & 0.1520 & 0.1580 & 0.1520 & 0.1260 & 0.1290 \\
 & 192 & 0.1673 & 0.1396 & 0.1880 & 0.1440 & 0.1680 & 0.1740 & 0.1720 & 0.1450 & 0.1470 \\
 & 336 & 0.1864 & 0.1535 & 0.2070 & 0.1620 & 0.1860 & 0.1910 & 0.2030 & 0.1620 & 0.1630 \\
 & 720 & 0.2295 & 0.1824 & 0.2560 & 0.1920 & 0.2280 & 0.2290 & 0.2890 & 0.1970 & 0.1970 \\
\bottomrule
\end{tabular}
\end{center}

\section{Reproducing the baseline renderer}
\label{app:repro-baseline}
For every LTSF cell, the frozen backbone at its published configuration (context length, fold period,
$r{=}c{=}0.4$) rendered three ways: by VisionTS's own column-wise renderer (run by us), by our row-wise
renderer (the value the zero-training tier uses as its anchor), and the number VisionTS publishes
(Table~17 of their paper, MAE-Base), which is what the main table's VisionTS-zs column reports. On ECL
VisionTS's own renderer at the published context reproduces the published row ($0.1772$ vs.\ $0.177$
at $H{=}96$) while our row-wise render of the same configuration is $9.4\%$ worse there, so the ECL gap
between the anchor and the published number is the orientation, not the context.
\begin{center}\scriptsize\begin{tabular}{llr|rrr}
\toprule
 & $H$ & context & VisionTS renderer (our run) & our row-wise render & published \\
\midrule
\multirow{4}{*}{ETTh1} & 96 & 2880 & 0.353 & 0.343 & 0.353 \\
 & 192 & 2880 & 0.392 & 0.369 & 0.392 \\
 & 336 & 2880 & 0.407 & 0.381 & 0.407 \\
 & 720 & 2880 & 0.406 & 0.409 & 0.406 \\
\midrule
\multirow{4}{*}{ETTh2} & 96 & 1728 & 0.271 & 0.273 & 0.271 \\
 & 192 & 1728 & 0.328 & 0.332 & 0.328 \\
 & 336 & 1728 & 0.345 & 0.354 & 0.345 \\
 & 720 & 1728 & 0.388 & 0.369 & 0.388 \\
\midrule
\multirow{4}{*}{ETTm1} & 96 & 2304 & 0.341 & 0.340 & 0.341 \\
 & 192 & 2304 & 0.360 & 0.362 & 0.360 \\
 & 336 & 2304 & 0.377 & 0.377 & 0.377 \\
 & 720 & 2304 & 0.416 & 0.423 & 0.416 \\
\midrule
\multirow{4}{*}{ETTm2} & 96 & 4032 & 0.228 & 0.230 & 0.228 \\
 & 192 & 4032 & 0.262 & 0.266 & 0.262 \\
 & 336 & 4032 & 0.293 & 0.299 & 0.293 \\
 & 720 & 4032 & 0.343 & 0.353 & 0.343 \\
\midrule
\multirow{4}{*}{Weather} & 96 & 4032 & 0.220 & 0.222 & 0.220 \\
 & 192 & 4032 & 0.244 & 0.244 & 0.244 \\
 & 336 & 4032 & 0.280 & 0.276 & 0.280 \\
 & 720 & 4032 & 0.330 & 0.320 & 0.330 \\
\midrule
\multirow{4}{*}{ECL} & 96 & 2880 & 0.177 & 0.194 & 0.177 \\
 & 192 & 2880 & 0.188 & 0.204 & 0.188 \\
 & 336 & 2880 & 0.207 & 0.224 & 0.207 \\
 & 720 & 2880 & 0.256 & 0.257 & 0.256 \\
\bottomrule
\end{tabular}
\end{center}

\section{The drifting long-horizon family}
\label{app:drift}
The trained tier's four losses are all long-horizon ETT cells (ETTh1-720 and ETTh2 at 192, 336 and
720), where the val$\to$test distribution shift flips which arms are skilled. Seven protocol-legal rescue detectors (val-internal transfer checks, input
drift gauges, rolling and fresh gates, \dots) were falsified: the quantity a validation-side
detector must key on is precisely the quantity the drift destroys. The fixed-share anchor is the
price of this undetectability --- it converts catastrophic losses into bounded ones at a
small tax on winning cells.

\section{Combining the arms: gate, equal weights, non-negative fit}
\label{app:combiner}
\begin{center}\includegraphics[width=.98\linewidth]{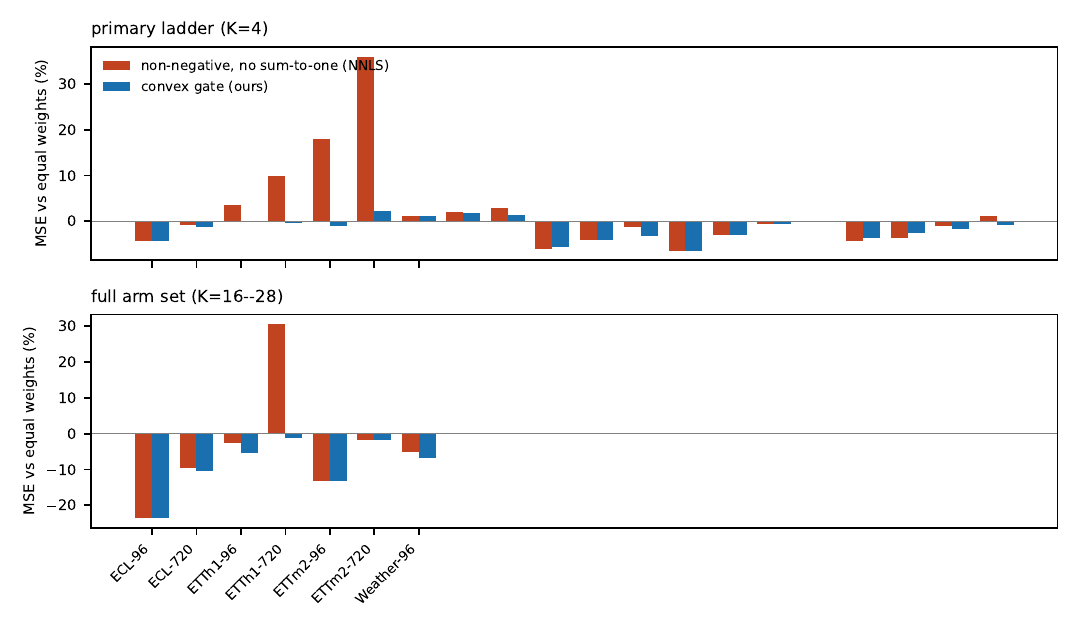}\end{center}
\noindent Per cell, MSE of non-negative least squares without the sum-to-one constraint (its weight
sums drift to 0.3--1.6) and of the convex gate relative to equal weights, on the primary ladder ($K{=}4$, top) and on the full arm set
(bottom). Dropping the sum-to-one constraint detonates on drifting cells (ETTh1-720 $+36\%$); the convex gate
never does, ties equal weights on homogeneous ladders and is what makes the heterogeneous full
set usable (ECL-96 $-23.5\%$ vs equal weights). Heterogeneity here is the spread of single-arm
error: on ECL the worst of the 16 arms, a half-period fold, has 4.6--6.7$\times$ the MSE of the best,
against at most 2.6$\times$ elsewhere.

\section{Sensitivity to the fusion share}
\label{app:lambda}
\begin{center}\includegraphics[width=.72\linewidth]{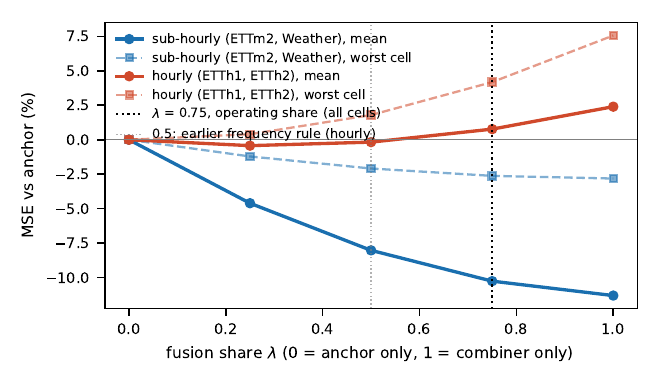}\end{center}
\noindent Zero-training tier, by family: change in MSE against the anchor-only point
($\lambda{=}0$) as the share moves toward the gate alone ($\lambda{=}1$). The operating
$\lambda{=}0.75$ sits on the plateau of both families: sub-hourly cells tolerate $\lambda{=}1$
(every cell improves, mean $-11.3\%$) while hourly cells need the anchor, their worst cell
degrading $+7.6\%$ there. Choosing $\lambda$ with test-set hindsight leaves little on the table:
per dataset it would gain $0.6\%$ on the trained tier's six-dataset average ($1.2\%$
zero-training), per cell $1.3\%$. ECL is the exception --- its anchor is weak, the published
render trailing the gated combination by 30--45\% --- so the earlier frequency rule
($\lambda{=}0.5$ on hourly data) costs $8$--$11\%$ per cell against the combination alone; at
$0.75$ the trained tier reads 0.150 (0.147 at $\lambda{=}1$, 0.161 at $0.5$) and the
zero-training tier 0.184 (0.183 / 0.191). The share cannot be chosen on validation: the gate is
fit on those windows, so validation prefers the combination on 22 of 24 cells, while the anchor's
value appears only under test-time drift.

\medskip\noindent\textbf{Anchor removed, anchor alone.} MSE relative to the operating
$\lambda{=}0.75$, by family, for the 24 trained-tier cells and the 16 cells of the zero-training
sweep (ILI excluded throughout):
\begin{center}\small\setlength{\tabcolsep}{3.5pt}\begin{tabular}{llr|rr|rr}
\toprule
 & & & \multicolumn{2}{c|}{without anchor ($\lambda{=}1$)} & \multicolumn{2}{c}{anchor only ($\lambda{=}0$)} \\
tier & family & cells & mean & worst cell & mean & worst cell \\
\midrule
trained & hourly/weekly & 12 & +0.4\% & +4.7\% (ETTh1-720) & +14.1\% & +40.6\% (ECL-720) \\
trained & sub-hourly & 12 & +0.6\% & +4.2\% (Weather-720) & +20.8\% & +54.2\% (Weather-96) \\
zero-training & hourly/weekly & 8 & +1.6\% & +3.5\% (ETTh1-720) & -0.8\% & +1.5\% (ETTh2-96) \\
zero-training & sub-hourly & 8 & -1.0\% & +0.5\% (Weather-720) & +10.3\% & +20.1\% (ETTm2-96) \\
\bottomrule
\end{tabular}
\end{center}
\noindent The two halves fail in opposite places. Without the anchor the hourly/weekly family
loses on its drifting cells (ETTh1-720 $+4.7\%$ trained, $+3.5\%$ zero-training) while the
sub-hourly family is indifferent. With the anchor alone the sub-hourly gains vanish ($+20.8\%$
trained, Weather-96 $+54\%$), and on hourly/weekly cells the cost is carried by ECL, whose
published render is far from optimal ($+40.6\%$ at ECL-720). The arms carry the gain where the
render is poor; the anchor bounds the loss where it is not.

\section{Arm axes: half-period folds and additional periods}
\label{app:multiperiod}
\begin{center}\includegraphics[width=.98\linewidth]{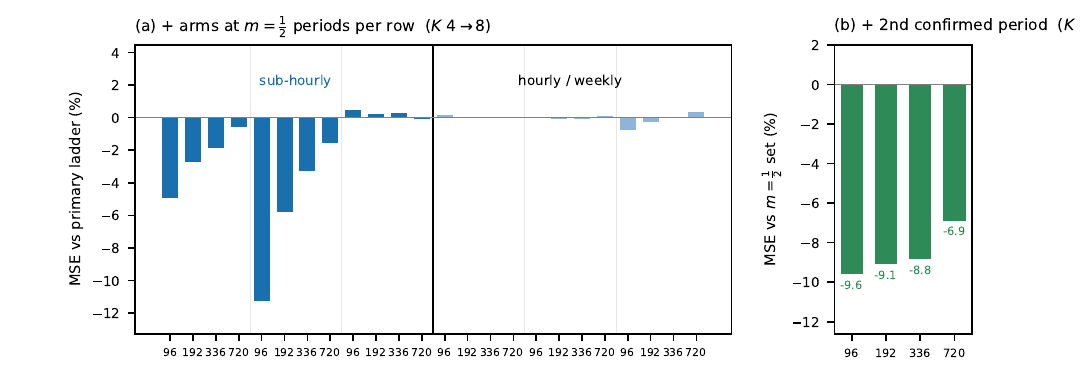}\end{center}
\noindent\textbf{Arm axes, one at a time.} Zero-training tier, the method's own arms at the operating
context, convex gate fit on validation, test MSE at the combination stage (no anchor). (a) adds the
$m{=}\tfrac12$ arms to the primary ladder ($K$ 4$\to$8) on all 24 cells; (b) adds the second confirmed
period ($K$ 8$\to$16), which on this benchmark exists only for ECL. Each step changes exactly one
generation axis, so each bar is attributable to one rule. Sub-hourly cells gain (ETTm2 and Weather
8/8, mean $-4.0\%$; ETTm1 neutral), hourly cells are unchanged (twelve cells within $\pm0.8\%$), and
ECL gains $-7$ to $-10\%$ from its second period on every horizon. The equal-weight columns of the same run show why the gate is a
prerequisite for the axis: with equal weights the $m{=}\tfrac12$ arms cost up to $+80\%$ on hourly
data (ECL-96 $0.175\!\to\!0.319$) while the gated combination is unchanged.

\medskip\noindent\textbf{On GIFT-Eval.}
The decomposition above --- primary ladder,
then the $m{=}\tfrac12$ arms, then every further confirmed period --- was re-run on the 30 configurations
whose detector confirmed and rendered a second period, with the zero-training protocol otherwise
unchanged. The
extra-period axis transfers: it lowers MASE on 19 of 30 configurations (geometric mean $-2.9\%$ at the
combination stage, $-2.6\%$ fused), and on the 18 where the gate gives those arms weight it helps on
16 (geometric mean $-5.0\%$; us\_births/W/short $-19.8\%$, m\_dense/H/short $-14.7\%$,
electricity/H/short $-11.5\%$). The $m{=}\tfrac12$ axis does not: it helps on 3 of 30 (geometric mean
$+2.0\%$). On the 18 configurations where the gate zeroes those arms the axis is free ($+0.3\%$); on
the 12 where it selects them from the four validation windows the benchmark provides, the selection
transfers on 2 (loop\_seattle/5T/short: 76\% of the gate weight on validation, $+33\%$ on test). On
LTSF, where the gate is fit on thousands of validation windows, the same axis
paid on every sub-hourly ETTm2 and Weather cell.

\section{Full per-configuration GIFT-Eval results}
\label{app:gift-master}
Both tiers under the final protocol (row-wise render, one seed), scored against the official
leaderboard result files. Periods, arms and gate follow the LTSF rules; the anchor folds at our
detected period rather than the baseline's per-dataset search, which is free at the aggregate
(0.899 either way on the anchor-only run). The only benchmark-specific settings are the two GIFT
leaves open: the validation windows and the per-window scale of the gate fit. The detector abstains
on 20 configurations, which fall back to the calendar period; on 3 more the context cannot hold two
cycles of the detected period, so no fold geometry exists and the zero-training tier emits the
anchor alone. The mean arm count is 10.0. Aggregates (Seasonal-Naive-normalised geometric means,
$\lambda{=}0.75$): zero-training 0.879 / 0.671, trained 0.832 / 0.634; the baseline column is its
leaderboard submission (0.863 / 0.755) while our own run of it, which Table~\ref{tab:gift}
compares against, scores 0.899 / 0.786. \textbf{Bold}: our MASE beats every listed baseline, our
CRPS beats the baseline's.

\paragraph{Predictive distribution for CRPS.} Every forecast is submitted as nine samples, the
Gaussian quantiles at levels 0.1 to 0.9 centred on the fused point, with per-position scale
$1.4826$ times the median absolute deviation of the arms and the anchor from the fused point; the
median sample is the fused forecast itself, so MASE is unaffected. Where the zero-training tier
emits the anchor alone (3 configurations) it submits a point. The baseline's leaderboard entry
submits a point forecast, whose CRPS equals its normalised absolute error (ND and CRPS coincide on
all 97 rows of its result file), and our re-run of it is scored the same way.

\paragraph{Numerical repair of the gate normaliser.} The gate divides each validation window's residual
by that window's context standard deviation and treated a window as uninformative only below an
absolute floor. A constant context at ordinary magnitude has a float32 standard deviation just
above that floor, so its residual entered the fit amplified by ${\sim}10^{5}$ and drove the weights
to zero, leaving the fused output at $(1-\lambda)$ times the anchor. The test is now relative to
the window's own magnitude, in float64. Six of the 97 configurations contain such a window (100
windows in all); the other 91 are unchanged by construction, and two controls reproduce their
stored values. Re-running the six in both tiers improves the three kdd\_cup\_2018/H configurations,
where the weights had collapsed, by 32--40\%, and moves the three bitbrains\_fast\_storage/5T
configurations, whose weights were well formed, by under 4\%. Before the repair the aggregates were
0.892 / 0.678 and 0.844 / 0.641.

\paragraph{Trained-tier arm sets.} Arms no series is long enough to train (context plus horizon
exceeds every series) are dropped; where that removes every arm the tier trains the single unfolded
default arm, the zero-training tier's own fallback (solar/W/short: 8 dropped). Otherwise the two
tiers share one arm set, the only difference being that the zero-training tier filters geometry per
context-length group while the trained tier uses the nominal context (car\_parts/M, m4\_weekly/W).

\medskip\noindent\textbf{Rule controls.} Two rules, each switched off on the configurations that
trigger it, the pipeline otherwise identical (GIFT-Eval, row-wise). \emph{Arm abstention}: when the
context cannot hold two cycles of the detected period the zero-training tier emits the anchor
alone, because the surviving $m{=}\tfrac12$ arms are views of a period the window never shows;
gating them instead costs up to $3.5\times$. After LayerNorm finetuning those same arms are
repaired and abstaining would forfeit up to $42\%$, so the trained tier does not abstain.
\begin{center}\small\setlength{\tabcolsep}{4pt}
\begin{tabular}{llrr}
\toprule
control & configuration & with rule & without \\
\midrule
arm abstention, zero-training & bizitobs\_l2c/5T/short & 0.443 & 1.564 \\
 & bizitobs\_l2c/5T/medium & 0.904 & 1.336 \\
 & bizitobs\_l2c/5T/long & 1.123 & 1.123 \\
arm abstention, trained tier & bizitobs\_l2c/5T/short & 0.452 & 0.439 \\
 & bizitobs\_l2c/5T/medium & 0.879 & 0.584 \\
 & bizitobs\_l2c/5T/long & 1.064 & 0.615 \\
anchor fold period & aggregate nMASE, 97 configs & 0.899 & 0.899 \\
\bottomrule
\end{tabular}
\end{center}
\begin{center}\tiny\setlength{\tabcolsep}{2pt}\renewcommand{\arraystretch}{0.9}\begin{tabular}{lrr|rr|rrrr|r}
\toprule
 & \multicolumn{2}{c|}{\method-zs} & \multicolumn{2}{c|}{\method (trained)} & \multicolumn{4}{c|}{MASE (official)} & CRPS \\
config & MASE & CRPS & MASE & CRPS & VisionTS & Chronos\textsubscript{b} & TimesFM & Naive & VTS \\
\midrule
bitbrains\_fast\_storage/5T/long & 1.115 & \textbf{0.949} & 1.190 & 1.006 & 1.260 & 0.948 & 0.980 & 1.137 & 0.968 \\
bitbrains\_fast\_storage/5T/medium & 1.186 & \textbf{0.856} & 1.304 & 0.981 & 1.260 & 1.062 & 1.075 & 1.220 & 0.897 \\
bitbrains\_fast\_storage/5T/short & 1.095 & 0.759 & 1.071 & 0.640 & 1.090 & 0.752 & 0.731 & 1.136 & 0.619 \\
bitbrains\_fast\_storage/H/short & 1.207 & \textbf{0.851} & 1.329 & \textbf{1.095} & 1.620 & 1.070 & 1.095 & 1.299 & 1.550 \\
bitbrains\_rnd/5T/long & 3.624 & \textbf{0.869} & 4.478 & \textbf{0.909} & 3.560 & 3.397 & 3.600 & 3.501 & 0.971 \\
bitbrains\_rnd/5T/medium & 4.629 & \textbf{0.772} & 6.020 & 0.938 & 4.620 & 4.449 & 4.595 & 4.542 & 0.908 \\
bitbrains\_rnd/5T/short & 2.127 & 0.657 & 2.122 & \textbf{0.566} & 2.090 & 1.705 & 1.769 & 1.971 & 0.650 \\
bitbrains\_rnd/H/short & 6.160 & \textbf{0.937} & 6.241 & 1.063 & 6.700 & 5.897 & 5.987 & 6.037 & 1.030 \\
bizitobs\_application/10S/long & 4.696 & \textbf{0.062} & 3.886 & \textbf{0.064} & 3.400 & 10.484 & 4.075 & 3.206 & 0.065 \\
bizitobs\_application/10S/medium & 4.259 & 0.058 & 3.234 & 0.055 & 2.720 & 9.720 & 3.082 & 2.691 & 0.052 \\
bizitobs\_application/10S/short & 2.319 & \textbf{0.036} & 2.087 & \textbf{0.037} & 2.250 & 5.533 & 1.563 & 2.242 & 0.042 \\
bizitobs\_l2c/5T/long & 1.123 & 0.683 & 0.634 & \textbf{0.313} & 0.608 & 1.241 & 1.236 & 1.454 & 0.347 \\
bizitobs\_l2c/5T/medium & 0.904 & 0.492 & \textbf{0.572} & \textbf{0.266} & 0.645 & 0.878 & 1.024 & 1.244 & 0.344 \\
bizitobs\_l2c/5T/short & 0.443 & 0.147 & 0.432 & 0.124 & 0.334 & 0.278 & 0.312 & 0.986 & 0.112 \\
bizitobs\_l2c/H/long & 0.771 & 0.419 & 0.713 & 0.381 & 0.635 & 0.556 & 1.299 & 1.426 & 0.371 \\
bizitobs\_l2c/H/medium & 0.552 & \textbf{0.296} & 0.621 & 0.319 & 0.503 & 0.495 & 1.183 & 1.510 & 0.302 \\
bizitobs\_l2c/H/short & 0.603 & \textbf{0.295} & 0.579 & \textbf{0.279} & 0.607 & 0.432 & 0.782 & 1.214 & 0.330 \\
bizitobs\_service/10S/long & 1.958 & 0.061 & 1.676 & \textbf{0.059} & 1.490 & 5.298 & 2.154 & 1.367 & 0.060 \\
bizitobs\_service/10S/medium & 2.129 & 0.056 & 1.551 & \textbf{0.049} & 1.320 & 4.977 & 1.531 & 1.321 & 0.049 \\
bizitobs\_service/10S/short & 1.301 & \textbf{0.034} & 1.117 & \textbf{0.025} & 1.210 & 3.316 & 1.037 & 1.225 & 0.041 \\
car\_parts/M/short & 1.238 & \textbf{1.362} & 1.148 & \textbf{1.263} & 1.120 & 0.855 & 0.922 & 1.201 & 1.470 \\
covid\_deaths/D/short & 44.335 & \textbf{0.057} & 40.744 & \textbf{0.045} & 45.400 & 38.855 & 47.366 & 46.912 & 0.058 \\
electricity/15T/long & 0.965 & 0.106 & 0.934 & \textbf{0.086} & 1.020 & 0.933 & 0.904 & 1.164 & 0.105 \\
electricity/15T/medium & 0.889 & \textbf{0.089} & 0.886 & \textbf{0.085} & 0.954 & 0.862 & 0.845 & 1.151 & 0.104 \\
electricity/15T/short & 1.367 & \textbf{0.124} & 1.205 & \textbf{0.110} & 1.540 & 0.935 & 0.907 & 1.717 & 0.168 \\
electricity/D/short & 1.624 & \textbf{0.071} & 1.760 & \textbf{0.072} & 1.690 & 1.448 & 1.486 & 1.987 & 0.091 \\
electricity/H/long & 1.353 & \textbf{0.109} & 1.338 & \textbf{0.113} & 1.390 & 1.238 & 1.055 & 1.524 & 0.127 \\
electricity/H/medium & 1.212 & \textbf{0.093} & 1.177 & \textbf{0.091} & 1.270 & 1.078 & 0.929 & 1.392 & 0.112 \\
electricity/H/short & 1.033 & \textbf{0.075} & 1.011 & \textbf{0.073} & 1.120 & 0.873 & 0.763 & 1.358 & 0.093 \\
electricity/W/short & 1.729 & \textbf{0.055} & 1.669 & \textbf{0.060} & 1.860 & 1.477 & 1.448 & 2.090 & 0.111 \\
ett1/15T/long & 1.098 & \textbf{0.270} & \textbf{1.062} & \textbf{0.279} & 1.080 & 1.136 & 1.114 & 1.191 & 0.320 \\
ett1/15T/medium & 1.087 & \textbf{0.275} & 1.070 & \textbf{0.287} & 1.270 & 1.061 & 1.082 & 1.188 & 0.370 \\
ett1/15T/short & 0.806 & \textbf{0.212} & 0.728 & \textbf{0.191} & 0.836 & 0.680 & 0.719 & 0.934 & 0.251 \\
ett1/D/short & 1.928 & \textbf{0.361} & 2.078 & \textbf{0.344} & 1.880 & 1.672 & 1.649 & 1.778 & 0.401 \\
ett1/H/long & 1.398 & \textbf{0.311} & 1.447 & \textbf{0.307} & 1.420 & 1.354 & 1.514 & 1.479 & 0.364 \\
ett1/H/medium & 1.430 & \textbf{0.297} & 1.439 & \textbf{0.314} & 1.400 & 1.375 & 1.310 & 1.568 & 0.372 \\
ett1/H/short & 0.861 & \textbf{0.201} & 0.841 & \textbf{0.205} & 0.850 & 0.828 & 0.866 & 0.977 & 0.241 \\
ett1/W/short & \textbf{1.634} & \textbf{0.340} & \textbf{1.616} & \textbf{0.347} & 1.730 & 1.697 & 1.650 & 1.769 & 0.395 \\
ett2/15T/long & 0.955 & \textbf{0.111} & 0.991 & \textbf{0.117} & 1.060 & 0.940 & 0.941 & 1.013 & 0.133 \\
ett2/15T/medium & 0.925 & \textbf{0.106} & 0.931 & \textbf{0.108} & 1.160 & 0.922 & 0.938 & 1.051 & 0.154 \\
ett2/15T/short & 0.915 & \textbf{0.089} & 0.784 & \textbf{0.077} & 0.952 & 0.766 & 0.747 & 1.067 & 0.102 \\
ett2/D/short & 1.858 & \textbf{0.120} & 1.983 & \textbf{0.124} & 1.660 & 1.322 & 1.559 & 1.390 & 0.137 \\
ett2/H/long & 1.099 & \textbf{0.121} & 1.126 & \textbf{0.122} & 1.290 & 1.036 & 1.126 & 1.128 & 0.162 \\
ett2/H/medium & 1.133 & \textbf{0.121} & 1.172 & \textbf{0.125} & 1.300 & 1.027 & 1.050 & 1.238 & 0.160 \\
ett2/H/short & 0.816 & \textbf{0.078} & 0.771 & \textbf{0.074} & 0.798 & 0.733 & 0.755 & 0.923 & 0.088 \\
ett2/W/short & 2.015 & 0.188 & 2.343 & 0.213 & 1.500 & 0.739 & 1.124 & 0.779 & 0.183 \\
hierarchical\_sales/D/short & 0.848 & \textbf{0.728} & 0.856 & \textbf{0.740} & 0.877 & 0.743 & 0.752 & 1.135 & 0.827 \\
hierarchical\_sales/W/short & 0.850 & \textbf{0.438} & 0.860 & \textbf{0.472} & 0.862 & 0.733 & 0.703 & 1.025 & 0.509 \\
hospital/M/short & 0.790 & \textbf{0.058} & 0.820 & \textbf{0.062} & 0.879 & 0.791 & 0.755 & 0.921 & 0.070 \\
jena\_weather/10T/long & 0.678 & \textbf{0.058} & 0.672 & \textbf{0.060} & 0.705 & 0.657 & 0.231 & 0.761 & 0.066 \\
jena\_weather/10T/medium & 0.661 & \textbf{0.059} & 0.653 & \textbf{0.061} & 0.663 & 0.610 & 0.191 & 0.716 & 0.068 \\
jena\_weather/10T/short & 0.514 & \textbf{0.044} & 0.421 & \textbf{0.041} & 0.397 & 0.306 & 0.091 & 0.743 & 0.061 \\
jena\_weather/D/short & 1.406 & \textbf{0.062} & 1.334 & \textbf{0.059} & 1.610 & 1.051 & 1.243 & 1.573 & 0.074 \\
jena\_weather/H/long & 1.160 & 0.081 & 1.132 & \textbf{0.075} & 1.100 & 1.028 & 1.057 & 1.268 & 0.081 \\
jena\_weather/H/medium & 0.916 & \textbf{0.066} & 0.868 & \textbf{0.063} & 0.841 & 0.747 & 0.864 & 0.889 & 0.068 \\
jena\_weather/H/short & 0.671 & \textbf{0.054} & 0.605 & \textbf{0.053} & 0.625 & 0.536 & 0.525 & 0.723 & 0.058 \\
kdd\_cup\_2018/D/short & \textbf{1.188} & \textbf{0.432} & \textbf{1.183} & \textbf{0.434} & 1.470 & 1.197 & 1.210 & 1.497 & 0.551 \\
kdd\_cup\_2018/H/long & 1.033 & \textbf{0.535} & 1.015 & \textbf{0.528} & 1.040 & 0.684 & 1.031 & 1.336 & 0.578 \\
kdd\_cup\_2018/H/medium & 1.078 & \textbf{0.504} & 1.067 & \textbf{0.503} & 1.060 & 0.700 & 1.026 & 1.429 & 0.525 \\
kdd\_cup\_2018/H/short & 1.035 & \textbf{0.458} & 0.973 & \textbf{0.446} & 1.200 & 0.601 & 0.941 & 1.340 & 0.609 \\
loop\_seattle/5T/long & \textbf{1.071} & \textbf{0.104} & \textbf{0.939} & \textbf{0.093} & 1.080 & 1.239 & 1.131 & 1.251 & 0.120 \\
loop\_seattle/5T/medium & \textbf{1.069} & \textbf{0.104} & \textbf{0.886} & \textbf{0.088} & 1.320 & 1.136 & 1.122 & 1.153 & 0.151 \\
loop\_seattle/5T/short & 0.757 & \textbf{0.067} & 0.596 & \textbf{0.057} & 0.790 & 0.628 & 0.583 & 0.762 & 0.087 \\
loop\_seattle/D/short & 0.918 & \textbf{0.049} & 0.935 & \textbf{0.050} & 0.950 & 0.903 & 0.859 & 1.732 & 0.056 \\
loop\_seattle/H/long & 1.077 & \textbf{0.079} & 0.966 & \textbf{0.072} & 1.020 & 0.996 & 0.906 & 1.546 & 0.088 \\
loop\_seattle/H/medium & 1.177 & \textbf{0.084} & 1.029 & \textbf{0.077} & 1.050 & 1.019 & 0.934 & 1.481 & 0.090 \\
loop\_seattle/H/short & 1.049 & \textbf{0.075} & 0.980 & \textbf{0.074} & 1.110 & 0.900 & 0.832 & 1.293 & 0.097 \\
m4\_daily/D/short & 5.345 & 0.039 & 4.648 & 0.034 & 4.010 & 3.198 & 3.086 & 3.278 & 0.033 \\
m4\_hourly/H/short & 0.927 & 0.046 & 0.870 & \textbf{0.027} & 0.925 & 0.837 & 0.596 & 1.193 & 0.039 \\
m4\_monthly/M/short & 1.112 & \textbf{0.116} & 1.046 & \textbf{0.110} & 1.160 & 0.949 & 0.600 & 1.260 & 0.139 \\
m4\_quarterly/Q/short & 1.382 & \textbf{0.094} & 1.229 & \textbf{0.088} & 1.360 & 1.224 & 0.965 & 1.602 & 0.104 \\
m4\_weekly/W/short & 4.546 & 0.060 & 4.062 & \textbf{0.053} & 2.530 & 2.078 & 2.222 & 2.777 & 0.057 \\
m4\_yearly/A/short & 3.646 & \textbf{0.137} & 3.227 & \textbf{0.121} & 3.830 & 3.507 & 2.538 & 3.966 & 0.159 \\
m\_dense/D/short & 0.822 & \textbf{0.082} & 0.965 & 0.100 & 0.788 & 0.716 & 0.636 & 1.669 & 0.094 \\
m\_dense/H/long & 0.976 & 0.170 & 0.805 & \textbf{0.139} & 0.821 & 0.938 & 0.795 & 1.478 & 0.166 \\
m\_dense/H/medium & 0.897 & \textbf{0.155} & 0.777 & \textbf{0.134} & 0.892 & 0.881 & 0.771 & 1.570 & 0.181 \\
m\_dense/H/short & 0.958 & \textbf{0.167} & 0.874 & \textbf{0.153} & 1.140 & 0.775 & 0.848 & 1.488 & 0.241 \\
restaurant/D/short & 0.727 & \textbf{0.317} & 0.697 & \textbf{0.305} & 0.743 & 0.700 & 0.692 & 1.006 & 0.358 \\
saugeen/D/short & 4.208 & \textbf{0.532} & 3.262 & \textbf{0.423} & 3.990 & 2.840 & 3.338 & 3.413 & 0.582 \\
saugeen/M/short & 0.774 & \textbf{0.345} & 0.770 & \textbf{0.345} & 0.800 & 0.739 & 0.836 & 0.976 & 0.410 \\
saugeen/W/short & 1.449 & \textbf{0.461} & 1.452 & \textbf{0.471} & 1.620 & 1.216 & 1.955 & 1.991 & 0.603 \\
solar/10T/long & \textbf{0.832} & \textbf{0.370} & 0.886 & \textbf{0.390} & 1.140 & 1.071 & 1.149 & 0.871 & 0.562 \\
solar/10T/medium & \textbf{0.858} & \textbf{0.394} & \textbf{0.861} & \textbf{0.387} & 0.890 & 1.027 & 1.169 & 0.927 & 0.451 \\
solar/10T/short & 1.246 & \textbf{0.682} & \textbf{0.962} & \textbf{0.554} & 1.110 & 0.991 & 1.478 & 1.106 & 0.736 \\
solar/D/short & 1.044 & \textbf{0.348} & 1.035 & \textbf{0.346} & 1.160 & 0.982 & 0.971 & 1.156 & 0.429 \\
solar/H/long & 0.992 & \textbf{0.385} & 0.966 & \textbf{0.379} & 0.956 & 1.033 & 1.266 & 1.071 & 0.435 \\
solar/H/medium & 1.036 & \textbf{0.404} & 0.988 & \textbf{0.396} & 0.889 & 0.931 & 0.959 & 0.935 & 0.418 \\
solar/H/short & 0.877 & \textbf{0.372} & 0.932 & \textbf{0.396} & 0.871 & 0.813 & 1.021 & 0.952 & 0.415 \\
solar/W/short & 1.939 & \textbf{0.359} & 1.574 & \textbf{0.273} & 2.140 & 0.980 & 1.282 & 1.470 & 0.395 \\
sz\_taxi/15T/long & 0.537 & \textbf{0.243} & 0.539 & \textbf{0.227} & 0.556 & 0.545 & 0.514 & 0.691 & 0.273 \\
sz\_taxi/15T/medium & 0.554 & \textbf{0.244} & 0.551 & \textbf{0.248} & 0.561 & 0.559 & 0.546 & 0.713 & 0.268 \\
sz\_taxi/15T/short & 0.592 & \textbf{0.246} & 0.568 & \textbf{0.247} & 0.626 & 0.548 & 0.539 & 0.764 & 0.292 \\
sz\_taxi/H/short & 0.584 & \textbf{0.161} & 0.572 & \textbf{0.161} & 0.594 & 0.562 & 0.560 & 0.738 & 0.182 \\
temperature\_rain/D/short & 1.687 & \textbf{0.737} & 1.811 & \textbf{0.754} & 1.600 & 1.304 & 1.433 & 2.012 & 0.781 \\
us\_births/D/short & 0.558 & \textbf{0.034} & 0.464 & \textbf{0.026} & 0.557 & 0.485 & 0.370 & 1.865 & 0.035 \\
us\_births/M/short & 0.734 & \textbf{0.017} & 0.545 & \textbf{0.013} & 0.727 & 0.924 & 0.497 & 0.761 & 0.020 \\
us\_births/W/short & 1.351 & \textbf{0.017} & 1.284 & \textbf{0.016} & 1.390 & 1.087 & 1.101 & 1.563 & 0.021 \\
\bottomrule
\end{tabular}
\end{center}

\end{document}